\documentclass[journal]{IEEEtran}
\usepackage{cite}
\ifCLASSINFOpdf
  \usepackage[pdftex]{graphicx}
\else
\fi
\usepackage{amsmath}
\usepackage{array}
\usepackage{url}
\usepackage{subcaption}
\usepackage[inline,shortlabels]{enumitem}
\usepackage{multirow}
\usepackage{booktabs}
\usepackage{soul,xcolor}
\usepackage[colorinlistoftodos,textsize=tiny]{todonotes}
\usepackage{xargs}
\usepackage{soul}
\usepackage{fancyhdr}

\newcolumntype{L}[1]{>{\raggedright\let\newline\\\arraybackslash\hspace{0pt}}m{#1}}
\newcolumntype{C}[1]{>{\centering\let\newline\\\arraybackslash\hspace{0pt}}m{#1}}
\newcolumntype{R}[1]{>{\raggedleft\let\newline\\\arraybackslash\hspace{0pt}}m{#1}}

\graphicspath {{images/}}
\setstcolor{red}
\fancypagestyle{copyright}
{
   \fancyhf{}
   \fancyfoot[C]{Copyright (c) 2017 IEEE. Personal use of this material is permitted. However, permission to use this material for any other purposes must be obtained from the IEEE by sending a request to pubs-permissions@ieee.org.}
}

\begin{document}

%
% paper title
% Titles are generally capitalized except for words such as a, an, and, as,
% at, but, by, for, in, nor, of, on, or, the, to and up, which are usually
% not capitalized unless they are the first or last word of the title.
% Linebreaks \\ can be used within to get better formatting as desired.
% Do not put math or special symbols in the title.
%\title{Joint Vertebrae Identification and Localization in Spinal CT Images by Combining 3D Convolutional and Recurrent Neural Networks}
\title{Joint Vertebrae Identification and Localization in Spinal CT Images by Combining Short- and Long-Range Contextual Information}
%it is not good to have CNN and RNN in the title :-)
%
%
% author names and IEEE memberships
% note positions of commas and nonbreaking spaces ( ~ ) LaTeX will not break
% a structure at a ~ so this keeps an author's name from being broken across
% two lines.
% use \thanks{} to gain access to the first footnote area
% a separate \thanks must be used for each paragraph as LaTeX2e's \thanks
% was not built to handle multiple paragraphs
%

\author{Haofu~Liao,~\IEEEmembership{Student Member,~IEEE,} Addisu Mesfin
        and~Jiebo~Luo,~\IEEEmembership{Fellow,~IEEE}% <-this % stops a space
\thanks{H. Liao and J. Luo is with the Department of Computer Science, University of Rochester, Rochester, NY, 14627 USA (e-mail: hliao6@cs.rochester.edu; jluo@cs.rochester.edu).}% <-this % stops a space
\thanks{A. Mesfin is with the Department of Orthopaedics, University of Rochester Medical Center, Rochester, NY, 14611 USA.}
}

% note the % following the last \IEEEmembership and also \thanks - 
% these prevent an unwanted space from occurring between the last author name
% and the end of the author line. i.e., if you had this:
% 
% \author{....lastname \thanks{...} \thanks{...} }
%                     ^------------^------------^----Do not want these spaces!
%
% a space would be appended to the last name and could cause every name on that
% line to be shifted left slightly. This is one of those "LaTeX things". For
% instance, "\textbf{A} \textbf{B}" will typeset as "A B" not "AB". To get
% "AB" then you have to do: "\textbf{A}\textbf{B}"
% \thanks is no different in this regard, so shield the last } of each \thanks
% that ends a line with a % and do not let a space in before the next \thanks.
% Spaces after \IEEEmembership other than the last one are OK (and needed) as
% you are supposed to have spaces between the names. For what it is worth,
% this is a minor point as most people would not even notice if the said evil
% space somehow managed to creep in.

% The paper headers
\markboth{}%
{LIAO \MakeLowercase{\textit{et al.}}: JOINT VERTEBRAE IDENTIFICATION AND LOCALIZATION IN SPINAL CT IMAGES BY COMBINING SHORT- AND LONG-RANGE CONTEXTUAL INFORMATION}
% The only time the second header will appear is for the odd numbered pages
% after the title page when using the twoside option.
% 
% *** Note that you probably will NOT want to include the author's ***
% *** name in the headers of peer review papers.                   ***
% You can use \ifCLASSOPTIONpeerreview for conditional compilation here if
% you desire.

% If you want to put a publisher's ID mark on the page you can do it like
% this:
%\IEEEpubid{0000--0000/00\$00.00~\copyright~2015 IEEE}
% Remember, if you use this you must call \IEEEpubidadjcol in the second
% column for its text to clear the IEEEpubid mark.

% use for special paper notices
%\IEEEspecialpapernotice{(Invited Paper)}

% make the title area
\maketitle

% As a general rule, do not put math, special symbols or citations
% in the abstract or keywords.
\begin{abstract}
Automatic vertebrae identification and localization from arbitrary CT images is challenging. Vertebrae usually share similar morphological appearance. Because of pathology and the arbitrary field-of-view of CT scans, one can hardly rely on the existence of some anchor vertebrae or parametric methods to model the appearance and shape. To solve the problem, we argue that
\begin{enumerate*}[1)]
\item one should make use of the \textit{short-range contextual information}, such as the presence of some nearby organs (if any), to roughly estimate the target vertebrae;
\item due to the unique anatomic structure of the spine column, vertebrae have fixed sequential order which provides the important \textit{long-range contextual information} to further calibrate the results.
\end{enumerate*}
We propose a robust and efficient vertebrae identification and localization system that can inherently learn to incorporate both the short-range and long-range contextual information in a supervised manner. To this end, we develop a multi-task 3D fully convolutional neural network (3D FCN) to effectively extract the short-range contextual information around the target vertebrae. For the long-range contextual information, we propose a multi-task bidirectional recurrent neural network (Bi-RNN) to encode the spatial and contextual information among the vertebrae of the visible spine column. We demonstrate the effectiveness of the proposed approach on a challenging dataset and the experimental results show that our approach outperforms the state-of-the-art methods by a significant margin.

\end{abstract}

% Note that keywords are not normally used for peerreview papers.
\begin{IEEEkeywords}
Automatic vertebrae identification and localization, CT image, deep learning, convolutional neural network, recurrent neural network, multi-task learning.
\end{IEEEkeywords}

% For peer review papers, you can put extra information on the cover
% page as needed:
% \ifCLASSOPTIONpeerreview
% \begin{center} \bfseries EDICS Category: 3-BBND \end{center}
% \fi
%
% For peerreview papers, this IEEEtran command inserts a page break and
% creates the second title. It will be ignored for other modes.
\IEEEpeerreviewmaketitle

\section{Introduction}

\IEEEPARstart{M}{edical} imaging techniques have been widely used in the diagnosis and treatment of spinal disorders. They provide physicians the essential tools for evaluating spinal pathologies and facilitate the spinal surgery by enabling noninvasive visualization for surgical planing and procedure. When evaluating spinal health, 3D imaging techniques, such as magnetic resonance imaging (MRI) and computed tomography (CT), are usually the first choices of healthcare providers as they give better views of the spinal anatomy. However,  identifying individual vertebra from 3D images, which is usually an initial step of reviewing and analyzing spinal images, is nontrivial and time-consuming \cite{Burns2015}.

Computational methods can be used to automate the quantitative analysis of spinal images and therefore enhance physicians' ability to provide spinal healthcare. In this context, we investigate the automation of localizing and identifying individual vertebrae from CT scans, which can substantially benefit the daily work of radiologists and many subsequent tasks in spinal image analysis. On one hand, the localization and identification results of individual vertebrae can be leveraged by many other computerized spinal analysis tasks, such as vertebral body/intervertebrae disc segmentation \cite{ben2012vertebral,alomari2015vertebral}, 3D spine reconstruction \cite{lecron2012fast,simons2014fast}, spinal image registration \cite{otake2012automatic,ruizrobust}, and so on. On the other hand, it may be a crucial component of many computer-aided diagnosis and intervention systems for spinal health \cite{yao2012detection,al2013compression,wang2016detection,linte2015toward,knez2016manual,kumar2015robotic}.

\begin{figure}[t]
\centering
\includegraphics[width=\linewidth]{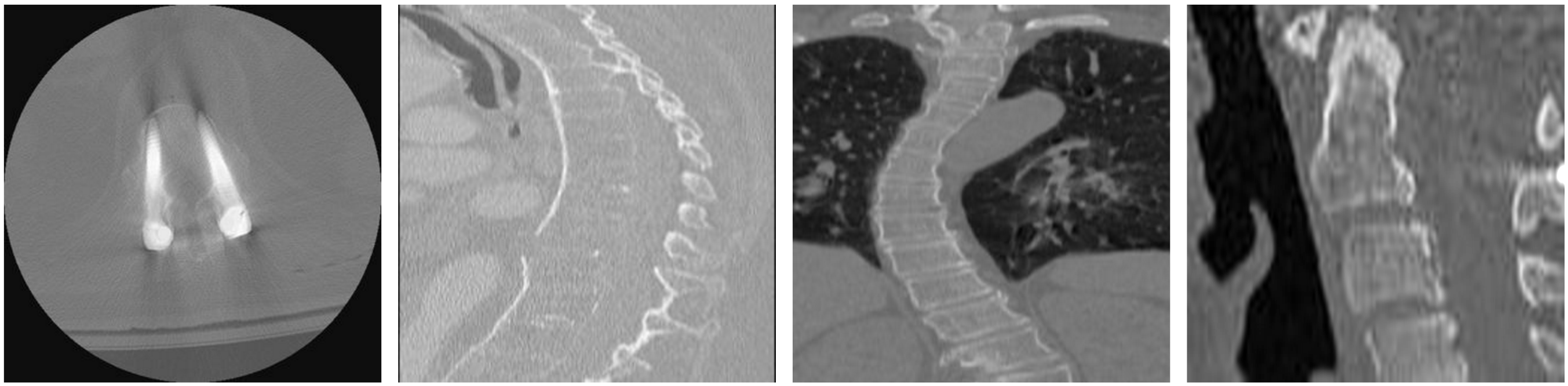}
\caption{The variability of spinal CT scans. Warping is performed for sagittal and coronal slices. Shown from left to right are the CT scan slices with surgical implants, blurry vertebrae boundaries, abnormal curvature and narrow field-of-view.}
\label{fig: spine_abnormal}
\end{figure}

However, designing a computerized vertebrae identification and localization system is nontrivial. Unlike other classification problems where the objects are often visually distinct, identifying individual vertebrae is challenging (demonstrated in Figure \ref{fig: spine_abnormal}) as neighboring vertebrae usually share similar morphological appearance. When the quality of the CT scan is low or only a narrow field-of-view is shown, it is really challenging to distinguish two neighboring vertebrae due to the similarity in  appearance. Moreover, because of pathologies, the anatomical structure of a vertebrae column is also not always regular and predictable and it gets even more complicated if a patient has surgical implants around the vertebrae, which often reduces the contrast of the vertebrae boundaries.

\begin{figure*}[t]
\centering
\includegraphics[width=0.8\linewidth]{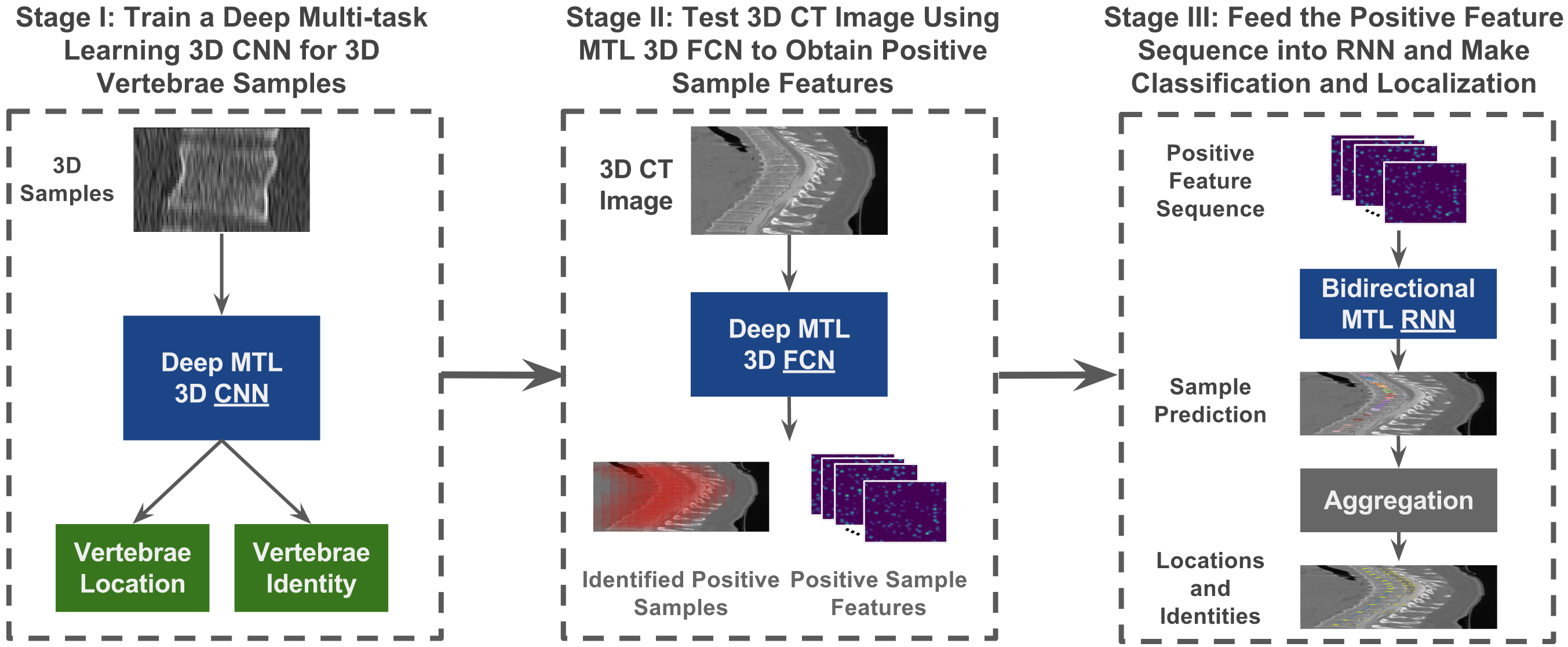}
\caption{Overall architecture of the proposed method for vertebrae identification and localization.}
\label{fig: spine_architecture}
\end{figure*}

Many methods have been proposed to identify and localize vertebrae automatically. Some early systems \cite{ma2010hierarchical,schmidt2007spine,kelm2010detection,zhan2012robust} usually require prior knowledge or have constraints about the content of the spinal images, making them less robust to more general cases in spinal imaging. In 2012, Glocker et al. \cite{glocker2012automatic} proposed a more general method that works for arbitrary field-of-view CT scans. However, their work makes assumptions about the shape and appearance of vertebrae, which may not be satisfied on pathological or abnormal spinal images. To address the limitations, Glocker et al. \cite{glocker2013vertebrae} further proposed a method that transforms sparse centroid annotations into dense probabilistic labels so that the modeling of shape and appearance can be avoided. However, these methods are based on handcrafted feature extraction methods which cannot encode more general visual characteristics of spinal images and as a result they fail to handle more complicated pathological cases when surgical implants exist. Chen et al. \cite{chen2015automatic} recently proposed to use convolutional neural networks (CNNs) to extract more robust features and their work achieved a superior performance on the same dataset as \cite{glocker2012automatic,glocker2013vertebrae}. In their work, they use 2D CNNs to encode the features of 3D CT volumes. Although it has been shown that for some segmentation tasks, applying 2D CNNs to 3D data can give reasonably well results as the segmentation itself can sometimes be addressed slice by slice which favors 2D operations \cite{DBLP:journals/corr/HarrisonXGLSM17,cai2016pancreas}. However, as denoted in \cite{dou2016automatic} and also demonstrated in this work, 2D CNNs do not work well in detection problems as they cannot capture the 3D spatial information that is critical to the detection of the target object. More recently, Dong et al. \cite{DBLP:journals/corr/YangXXHLZXPCTCM17} proposed a 3D U-Net \cite{ronneberger2015u} like architecture to target the vertebrae localization problem in an image-to-image fashion. However, the proposed architecture can not fully address the long-term contextual information in spinal images. To compensate this limitation, they further introduce a message passing and sparsity regularization algorithm for refinement.
Although the state-of-the-art methods have already achieved acceptable performance on a challenging 3D spine dataset \cite{glocker2012automatic,glocker2013vertebrae}, we argue that to further improve the vertebrae identification and localization performance, a computerized system should
\begin{enumerate*}[(1)]
  \item use a 3D feature extraction scheme such that it can better leverage the \textit{short-range contextual information} e.g., the presence of nearby organs of the target vertebrae;
  \item process the 3D spinal image in a sequential manner with the ability of encoding the \textit{long-range contextual information} e.g., the fixed spatial order of the vertebrae;
  \item learn the vertebrae identification and localization simultaneously and share the domain information of the two tasks during the training.
\end{enumerate*}

\begin{figure*}[t]
\centering
\includegraphics[width=0.8\linewidth]{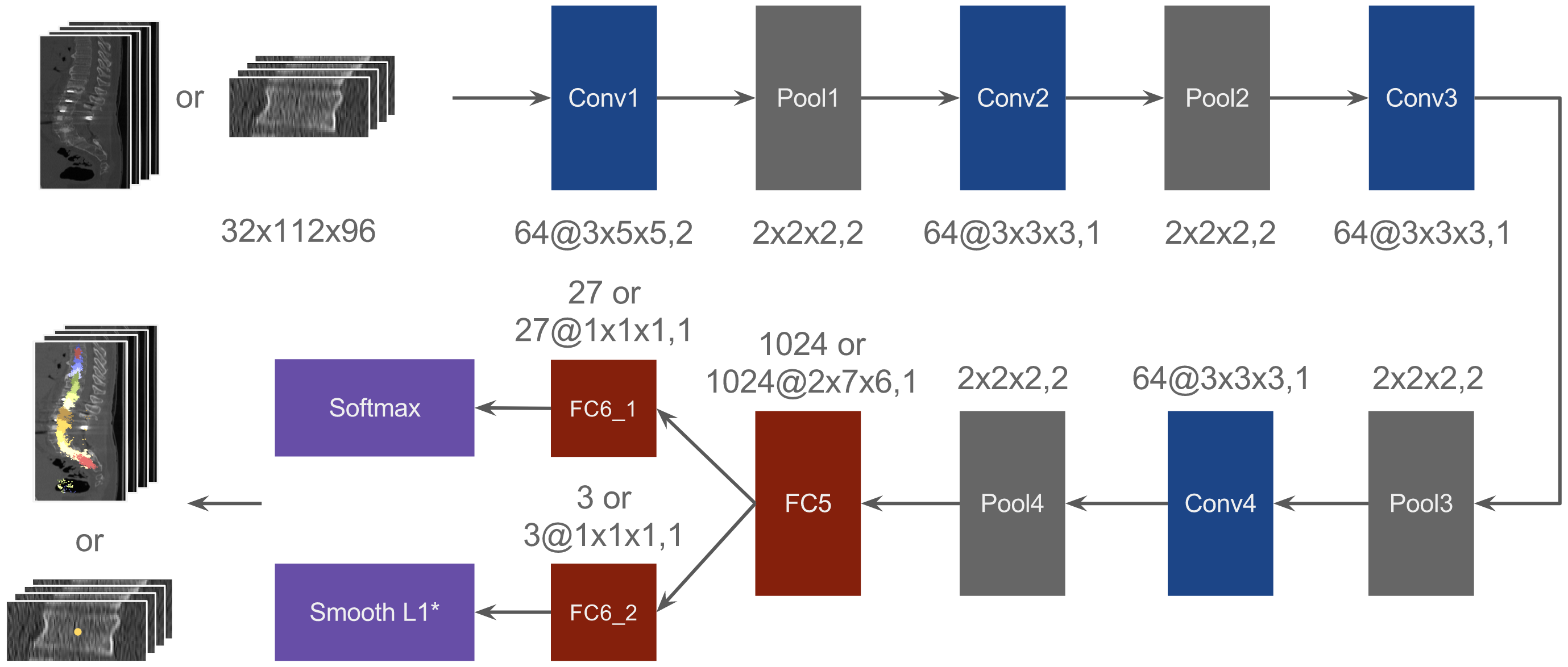}
\caption{The architecture of the deep multi-task 3D CNN/FCN. Note for CNN the input is a cropped image sample of size $32\times112\times196$ and the last two layers FC5 and FC6(\_1 or \_2) are fully connected layers. For FCN the input is a CT image of any size and the last two layers are convolutional layers.}
\label{fig: spine_cnn}
\end{figure*}

To this end, we propose a novel method which jointly learns vertebrae identification and localization by combining 3D convolutional and recurrent neural networks. We first develop a 3D fully convolutional neural network (FCN) to extract features of CT images in a sliding window fashion. The proposed 3D FCN employs 3D convolutional layers as its core components. 3D convolutional layers can encode 3D contextual information of the receptive field, which gives a better feature representation of the 3D spinal image than their 2D counterparts. To further improve the feature extraction, the proposed 3D FCN is trained in a multi-task learning (MTL) manner \cite{girshick2015fast} that leverages both the vertebrae centroid and name information simultaneously. The extracted features of the spinal images, however, only encode the short-range contextual information of each sampling area. Due to the special anatomical structure of spine column, vertebrae in spinal images have a fixed spatial order which provides important long-range contextual information. To incorporate this domain-specific information into our model, we further propose to use a recurrent neural network (RNN) to encode the long-range contextual information that persists in the spinal images. Specifically, we develop a bidirectional MTL RNNs to jointly learn the long-range contextual information from two directions (from cervical vertebrae to sacral vertebrae and the other way around) for both the identification and localization tasks.

Note we are not the first to introduce RNN to solve medical related problems. RNN-enabled architectures have been used by existing works for key frame detection from medical videos \cite{chen2015automaticfetal,kong2016recognizing} or for biomedical image segmentation \cite{chen2016combining,DBLP:journals/corr/CaiLXXY17}. Video-based problems are handled at \textit{frame-level} with conventional CNNs and RNNs are used to capture the temporal relations between frames. Segmentation-based problems are solved at \textit{pixel-level} by image-to-image networks (e.g., U-Net or FCN) and RNNs are used to refine the results. While, in this context, we focus on regressing the vertebrae locations as well as predicting their corresponding types. Neither of the existing RNN-related approaches can directly address this task. Instead, we argue that for our problem CT images should be processed at \textit{sample-level} for location regression and vertebrae classification. Hence, we propose a novel framework which first uses a 3D FCN to jointly scan the CT images for vertebrae locations and types at sample-level and then uses an RNN to capture the structural relations between samples. To facilitate the unique sample-level approach, the FCN is developed through a two-stage design: in the first stage a 3D CNN is trained using CT image samples and in the second stage the trained CNN is converted to FCN for fast sample scanning from CT images.  The RNN accordingly is deployed to adapt the sample feature sequences. As a result, the formulated approach can incorporate both the short- and long-range information for better structural understanding other than learning the spatial-temporal relations or the pixel-level contextual information.

% We also note that an RNN-based approach has also been proposed for vertebrae localization and identification \cite{yang2017deep} recently. However, it is formulated as a segmentation problem and closely follows the pixel-level approaches.

The contributions of this paper are summarized as follows:
\begin{itemize}
\item We propose a novel multi-task 3D CNN for landmark detection. The proposed architecture encodes better feature representations by jointly learning classification and regression with 3D convolutions, which can benefit many landmark detection problems that use 3D medical images.
\item We improve the general 3D FCN framework by introducing RNNs to incorporate the long-range contextual information in 3D images. This RNN based approach can be useful for many other similar problems in 3D medical image processing where the target objects usually have similar anatomic structures and thus contextual information is critical.
\item The proposed approach outperforms the state of the art on a challenging dataset by a significant margin.
\end{itemize}

\section{Methodology} \label{sec: methodology}

The overall architecture of the proposed method is illustrated in Figure \ref{fig: spine_architecture}. We use a three-stage approach to solve the problem. In the first stage, a deep multi-task 3D CNN is trained using randomly cropped 3D vertebrae samples. The idea of using 3D convolutional layers for medical images as well as cropping 3D samples for training is inspired by Dou et al. \cite{dou2016automatic}. Compared with other deep learning approaches \cite{suzani2015fast,chen2015automatic} where only 2D convolutional layers were used, using 3D CNN retains the 3D spatial information of the input and encodes better feature representations. To learn a better model, the identification and localization tasks are trained simultaneously through MTL. In the second stage, we transform the trained multi-task 3D CNN into a multi-task 3D FCN by converting the fully connected layers to 3D convolutional layers. 3D FCN can be efficiently applied to 3D images of any size and produce a prediction map for the effective 3D samples. This idea is adapted from \cite{sermanet2013overfeat,long2015fully} and we use it to extract the features of all the positive samples of the input 3D image. Finally, in the third stage, the extracted sample features will be ordered and form a set of feature sequences. Those spatially ordered sample features will be used to train a bidirectional RNN (Bi-RNN) \cite{schuster1997bidirectional} that predicts the vertebrae locations and identities in the testing phase. The final results will then be generated via aggregation.

\subsection{Stage I: Deep Multi-Task 3D CNN}

In this stage, we aim to train a network that takes a relatively small and fixed-size ($32 \times 112 \times 96$ in this paper) 3D sample as the input and predicts the most likely vertebrae type and the corresponding centroid location. Note a sample may contain more than one vertebra and the network only predicts the one that is closest to the sample centre. When applied to a CT image (see Section \ref{sec: spine_fcn}), this network can be used to effectively exploit the short-range contextual information.

As shown in Figure \ref{fig: spine_cnn}, the proposed CNN has four convolutional layers, four pooling layers and three fully-connected layers. For convolutional layers, we pad the inputs such that the outputs from the layers have the same size as the inputs. For pooling layers, no padding is performed as we want to downsize the inputs for dense feature representation. The numbers associated with each convolutional layer denote the feature size, kernel size and stride size, respectively. The numbers above each fully-connected layer denote the output sizes. The feature and kernel sizes are chosen empirically with reference from \cite{krizhevsky2012imagenet,dou2016automatic,he2016deep}. The ``FC5'' layer serves as the feature layer that encodes the final features for each input image sample. ``FC6\_1'' and ``FC6\_2'' layers serve as the prediction layer for vertebrae identification and localization, respectively. The output size of ``FC6\_1'' is $27$ as we have $26$ different vertebrae types plus the background and the output size of ``FC6\_2'' is $3$ because the location has $3$ dimensions. The input sample size is $32 \times 112 \times 96$. We choose this size based on several considerations: 
\begin{enumerate*}[(1)]
  \item this size should cover most of the vertebrae in the training set;
  \item each dimension should be a multiple of $16$ such that the feature map sizes are still integers after 4 pooling layers;
  \item the ratio of the three dimensions should approximate the shape of vertebrae.
\end{enumerate*}

The proposed CNN is trained using randomly cropped vertebrae samples. In particular, we call the samples that contain at least one vertebrae centroid \textit{positive samples} and use the label of the closest vertebrae centroid (to the sample centre) as the sample label. For those samples that do not contain any vertebrae centroids, we call them \textit{negative samples} and assign the background label to those samples. In total, there are $26$ vertebrae types with labels from C1-C7, T1-T12, L1-L5 and S1-S2. For convenience, we assign each of the label an integer with $\text{C1}=0$, $\dots$, $\text{S1}=25$ and $\text{background}=26$.

\begin{figure}[t]
\centering
\includegraphics[width=\linewidth]{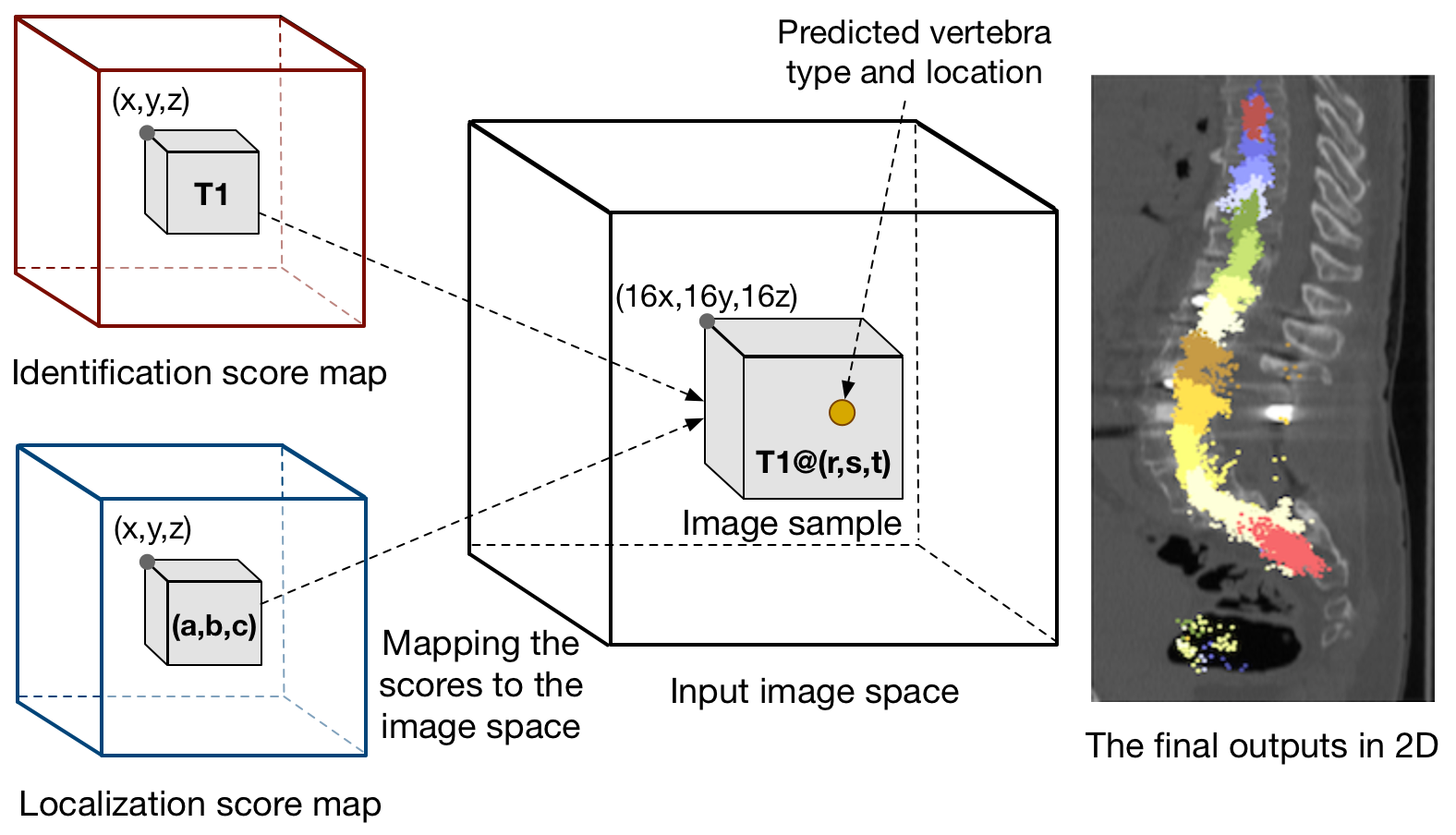}
\caption{3D score maps and their mapping to the image space.}
\label{fig: spine_mapping}
\end{figure}

To jointly learn vertebrae identification and localization, two losses are used for each of the tasks, respectively. The total loss is given by
\begin{equation}
L = L_{\text{id}} + \lambda L_{\text{loc}}
\end{equation}
where $L_{\text{id}}$ denotes the identification loss, $L_{\text{loc}}$ denotes the localization loss and $\lambda$ denotes the importance coefficient that controls the relative learning rate of the two tasks. We use a cross entropy softmax loss, which is commonly used for classification problems, for the identification task. Let $\{\mathbf{x}_0, \mathbf{x}_1, \dots, \mathbf{x}_{N-1}\}$ be a set of $N$ image samples and $\{\mathbf{y}_0, \mathbf{y}_1, \dots, \mathbf{y}_{N-1}\}$ be a set of $N$ ground truth labels where each label is a one-hot vector denoted as $\mathbf{y_i} = [y_{i0}, y_{i1}, \dots, y_{iP-1}]^T, y_{ij} \in \{0, 1\}, i \in \{0, 1, \dots, N-1\}, j \in \{0, 1, \dots, P-1\}$. The identification loss $L_{\text{id}}$ can be written as
\begin{multline}
L_{\text{id}} = -\frac{1}{N}\sum_{i=0}^{N-1}\sum_{j=0}^{P-1}y_{ij}\log{(f_{\text{id}}^j(\mathbf{x}_i;\mathbf{W}))} + \\ (1 - y_{ij})\log{(1 - f_{\text{id}}^j(\mathbf{x}_i;\mathbf{W}))}
\end{multline}
where $f_{\text{id}}^j$ denotes $j$-th output of the ``FC6\_1'' layer and $\mathbf{W}$ denotes all the network parameters. For the localization task, it is a regression problem. Therefore, we use a smooth L1 loss \cite{girshick2015fast} for this task. Given a set of $N$ ground truth locations $\{\mathbf{p}_0, \mathbf{p}_1, \dots, \mathbf{p}_{N-1}\}$ where $\mathbf{p}_i = [p_{i0}, p_{i1}, \dots, p_{iD-1}]^T, i \in \{0, 1, \dots, N-1\}$, the localization loss $L_{\text{loc}}$ can be written as
\begin{equation}
L_{\text{loc}} = \frac{1}{m}\sum_{i=0}^{N-1}[y_{iP-1}=0]\sum_{j=0}^{D-1}\text{smooth}_{L_1}(p_{ij}-f_{\text{loc}}^j(\mathbf{x}_i;\mathbf{W}))
\end{equation}
where $m = \sum_{i=0}^{N-1}[y_{iP-1}=0]$, $f_{\text{loc}}^j$ is the $j$-th output of the ``FC6\_2'' layer and $\text{smooth}_{L_1}(x)$ is the smooth L1 loss that is given by
\begin{equation}
\text{smooth}_{L_1}(x) =
\begin{cases}
0.5x^2, &\text{if } |x| < 1, \\
|x| - 0.5, &\text{otherwise}.
\end{cases}
\end{equation}
Note that only the locations for positive samples are meaningful. Thus, the localization loss will only be computed for positive samples and for negative samples the localization loss is zero. Here, we use the Iverson bracket indicator function $[y_{iP-1}=0]$ to ignore negative samples. $P-1$ indicates the background label and $y_{iP-1}=0$ means that the one-hot vector corresponds to a non-background label.

\subsection{Stage II: Deep Multi-Task 3D FCN} \label{sec: spine_fcn}

We use the trained deep multi-task 3D CNN to encode the short-range contextual information in a CT scan image. A straightforward approach is scanning the image in a sliding window manner by repeatedly cropping and processing overlapped image samples. When the input image is large, this approach can be very expensive and inefficient. As a solution, we propose to transform the CNN into a FCN. FCNs only contain convolutional and pooling layers. As pooling and convolution operation are computed using sliding windows, FCNs essentially process the input images in a sliding window manner but through the more efficient pooling and convolution operation.

When converting a CNN to a FCN, we must make sure the output of a FCN is identical to the output of a CNN if the input image size of FCN is the same as the input size required by CNN. This requires the convolutional layers of FCN should share the same weights and kernel layout as the corresponding fully connected layers in CNN. As denoted in Figure \ref{fig: spine_cnn}, for the ``FC5'' layer, we convert it to a convolutional layer with parameters configured as ``$1024@2 \times 7 \times 6, 1$''. We use a kernel size of $2 \times 7 \times 6$ because the output feature map size from ``Pool4'' is $2 \times 7 \times 6$ when the input image size is the same as the sample size, i.e., $32 \times 112 \times 96$. Similarly, ``FC6\_1'' and ``FC6\_2'' are converted to convolutional layers with configurations of ``$27@1 \times 1 \times 1, 1$'' and ``$3@1 \times 1 \times 1, 1$'', respectively. After this conversion, the constructed FCN will have the same number of parameters and kernel layouts as the trained CNN. Thus, we use the trained parameters from CNN to initialize the FCN which, as a result, gives the same outputs as the ones from CNN.

\begin{figure}[t]
\centering
\includegraphics[width=\linewidth]{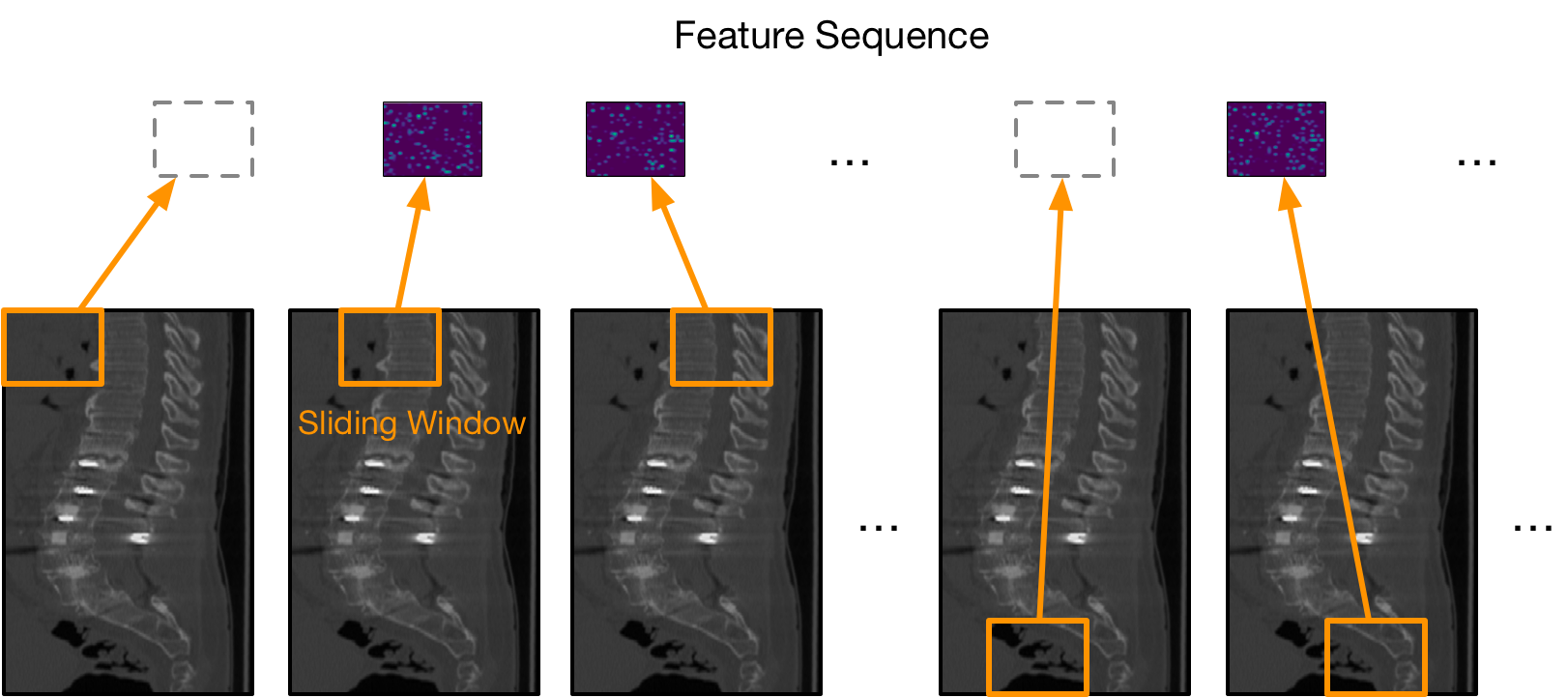}
\caption{Feature sequence generation using the proposed FCN.}
\label{fig: spine_features}
\end{figure}

\begin{figure*}[t]
\centering
\includegraphics[width=0.9\linewidth]{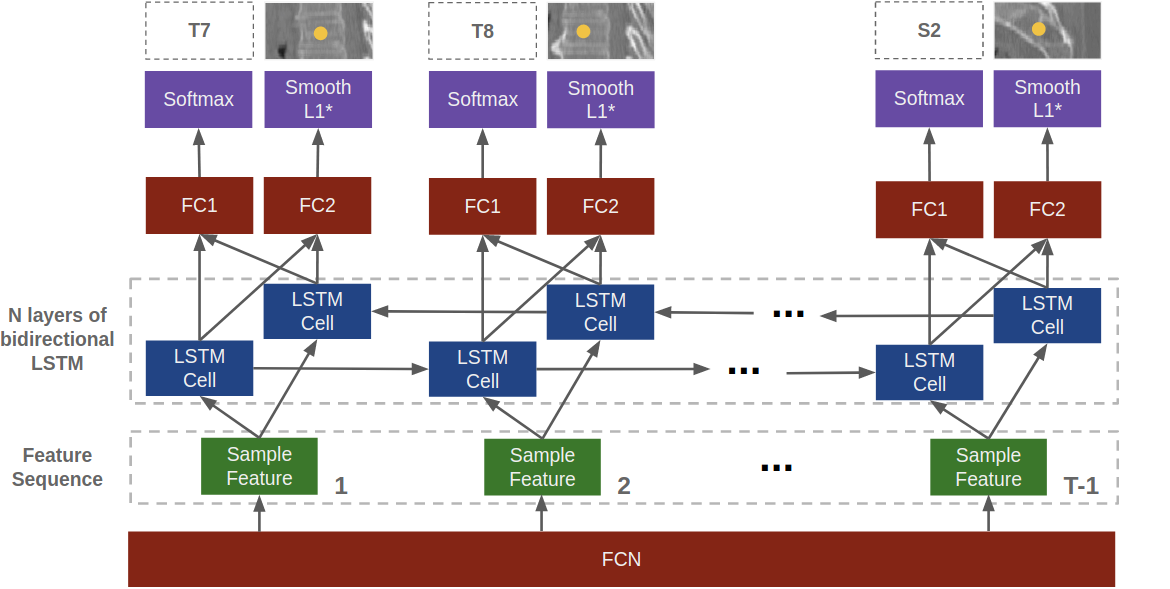}
\caption{The architecture of the multi-task Bi-RNN (unrolled in time).}
\label{fig: spine_rnn}
\end{figure*}
As shown in Figure \ref{fig: spine_mapping}, the FCN outputs two 3D score maps \footnote{Strictly speaking, the maps are 4D as each score itself is a vector.}, one for identification and the other for localization. Each score is a vector indicating either the vertebrae label or the vertebrae location. The scores at the same location of the two maps can be mapped to the same image sample in the CT image space. As there are 4 pooling layers in the proposed FCN with each has a stride of $2$, the effective sliding window stride is $2^4=16$. Therefore, for a score at $p_{score} = (x, y, z)$ of the output 3D score map, the corresponding image sample location at the input image is given by $p_{sample} = (16x, 16y, 16z)$. The identification score decides the vertebrae type of the image sample and the localization score gives the vertebrae centroid location in the sample. Assuming the predicted centroid location is $c_{score}=(a, b, c)$, then the corresponding location at the image space is given by $c_{sample}(r,s,t)=(16x + a, 16y + b, 16z + c)$. For each of the predicted centroid location, we assign it with the vertebrae label from the corresponding identification score. The final results are a set of densely predicted centroid points, as demonstrated in Figure \ref{fig: spine_mapping}, with each color indicating a different vertebrae type.

The proposed FCN can be used to extract feature sequence from the CT image, which will be further used to feed the RNN in the next stage (see Section \ref{sec: spine_rnn}). Figure \ref{fig: spine_features} illustrates the process of generating the feature sequence. Given an input image, the FCN implicitly processes it in a sliding window fashion. In addition to the 3D score maps, we can also obtain the feature maps from the intermediate layers and in this paper we extract the high-level features from the ``FC\_5'' layer \footnote{The feature map from this layer is a vector. We resize it to 2D in Figure \ref{fig: spine_mapping} for visualization purpose.}. Since for each sample image we can obtain a feature map, the final output is a sequence of feature maps. Note we ignore the feature maps of background sample images as they contain no vertebrae and have little contribution to the vertebrae identification and localization.

\subsection{Stage III: Multi-Task Bidirectional RNN} \label{sec: spine_rnn}

FCN can be used directly for vertebrae identification and localization by aggregating the predication results of all the positive samples. However, the limitation is that FCN can only encode short-range contextual information inside each sample image. But for vertebrae identification, long-range contextual information of neighboring vertebrae is also very helpful. For example, the algorithm may have difficulty distinguishing a vertebra between T2 and T3. However, if it knows that the vertebra above current vertebra is very likely to be T1 then it would be more confident to classify the current vertebra as T2 than T3. To this end, we propose to use RNN to encode the long-range contextual information between vertebrae. The idea is first converting a CT scan image into a sequence of spatially ordered vertebrae sample features using FCN and then feeding the sequence into an RNN that has already learned to encode the long-range contextual information among samples.

The architecture of the proposed RNN is illustrated in Figure \ref{fig: spine_rnn}. Here, we use a Bi-RNN structure instead of a conventional one. Such a choice is based on the observation that the contextual information of current vertebra may come from two directions: the vertebrae above and below. The RNN cell used in the network is long-short term memory (LSTM) \cite{hochreiter1997long} cell which can handle long sequences. Each LSTM cell has $256$ hidden states. As denoted in Figure \ref{fig: spine_rnn}, a bidirectional LSTM layer consists of two LSTM cells of opposite directions. The outputs of the two LSTM cells are concatenated together to form the final output of the bidirectional LSTM layer. We stack $N$ layers of such bidirectional LSTM to encode the input feature sequence. In our experiments $N=3$ and we find minor performance difference with more bidirectional LSTM layers. Two fully-connected layers ``FC1'' and ``FC2'' are put at the top of the bidirectional LSTM layers for computing the classification and regression scores, respectively. The input at each timestep is a sample feature vector extracted from FCN as illustrated in Figure \ref{fig: spine_features}. The RNN is also trained in a multi-task learning manner. At the end of each timestep, there are two loss functions, one for the identification loss and the other for the localization loss. We use the same loss functions as the ones used in Stage I, but the total loss is accumulated over time
\begin{equation}
L = \sum_{t=0}^{T-1}L_{\text{id}}^t + \lambda L_{\text{loc}}^t
\end{equation}
To train the Bi-RNN, we need to generate a set of sample feature sequences from the training set. To increase the data variation, we first augment the training set by randomly cropping subimages (not the fixed-size samples) from the training CT scans. For each of the cropped subimages, we feed it through the FCN that samples the input image in a sliding window manner. Next, the features of the samples that are labeled as positive are kept and ordered based on the samples' relative spatial locations. Each of the spatially ordered sample feature sequences will be used as the training data for the Bi-RNN. Note the associated vertebrae labels and centroid points for each of the sample features can be easily calculated from the ground truth annotations that come with the CT scan images.

During the testing phase, each of the new CT scan image will first be processed by the FCN to generate the ordered sample feature sequence. Next, the sample feature sequence will be passed to the Bi-RNN which outputs the identification and localization results for each of the samples at each timestep. Note they are dense identification and localization results as the samples are overlapped. Finally, we aggregate the dense results using the median \footnote{We use median instead of mean to suppress the outliers.} of the localization results that have the same identification label.

\begin{table}[]
\centering
\caption{Ablation study of the proposed multi-task 3D CNN}
\label{tab: spine_exp_cnn}
\begin{tabular}{@{}lcc@{}}
\toprule
           & \textbf{Cls. Accuracy} & \textbf{Loc. Error (mm)} \\ \midrule
2D CNN MTL & 43.19\%               & 8.74                     \\
3D CNN ID & 48.52\%                & N/A                      \\
3D CNN LOC & N/A                   & 7.05                     \\
3D CNN MTL & \textbf{52.39\%}      & \textbf{7.03}            \\ \bottomrule
\end{tabular}
\end{table}

\section{Experiments}

\begin{table*}[t]
\centering
\caption{Comparison of the proposed method with the state-of-the-art methods}
\label{tab: spine_exp_rnn}
\begin{tabular}{@{}cccccccccccccccccccccccc@{}}
\toprule
         & \multicolumn{3}{c}{\textbf{Glocker et al.}\cite{glocker2013vertebrae}} &  & \multicolumn{3}{c}{\textbf{Chen et al.}\cite{chen2015automatic}} &  & \multicolumn{3}{c}{\textbf{Dong et al.}\cite{DBLP:journals/corr/YangXXHLZXPCTCM17}} &  & \multicolumn{3}{c}{\textbf{Ours CNN}} &  & \multicolumn{3}{c}{\textbf{Ours CNN+RNN}}       &  & \multicolumn{3}{c}{\textbf{Ours CNN+Bi-RNN}}    \\ \midrule
Region   & Id. Rate         & Mean          & Std           &  & Id. Rate        & Mean         & Std          &  & Id. Rate      & Mean      & Std              &  & Id. Rate      & Mean      & Std       &  & Id. Rate        & Mean          & Std           &  & Id. Rate        & Mean          & Std           \\
All      & 74.0\%           & 13.20         & 17.83         &  & 84.2\%          & 8.82         & 13.04        &  & 85\%          & 8.6       & \textbf{7.8}     &  & 83.8\%        & 9.07     & 10.16     &  & 87.4\% & 6.59 & 8.71          &  & \textbf{88.3\%} & \textbf{6.47} & 8.56          \\
Cervical & 88.8\%           & 6.81          & 10.02         &  & 91.8\%          & 5.12         & 8.22         &  & 92\%          & 5.6       & \textbf{4.0}     &  & 91.2\%        & 8.63     & 11.17     &  & 93.8\% & 4.99 & 5.53 &  & \textbf{95.1\%} & \textbf{4.48} & 4.56 \\
Thoracic & 61.8\%           & 17.35         & 22.3          &  & 76.4\%          & 11.39        & 16.48        &  & 81\%          & 9.2       & \textbf{7.9}     &  & 79.1\%        & 9.56     & 10.08     &  & 81.7\% & 8.03 & 10.26          &  & \textbf{84.0\%} & \textbf{7.78} & 10.17          \\
Lumbar   & 79.9\%           & 13.05         & 12.45         &  & 88.1\%          & 8.42         & 8.62         &  & 83\%          & 11.0      & 10.8             &  & 85.1\%        & 8.57     & 8.87      &  & 89.4\% & 6.15 & 8.29 &  & \textbf{92.2\%} & \textbf{5.61} & \textbf{7.68} \\ \bottomrule
\end{tabular}
\end{table*}

\begin{figure*}[t]
\begin{subfigure}[b]{0.5\textwidth}
\centering
\includegraphics[width=\linewidth]{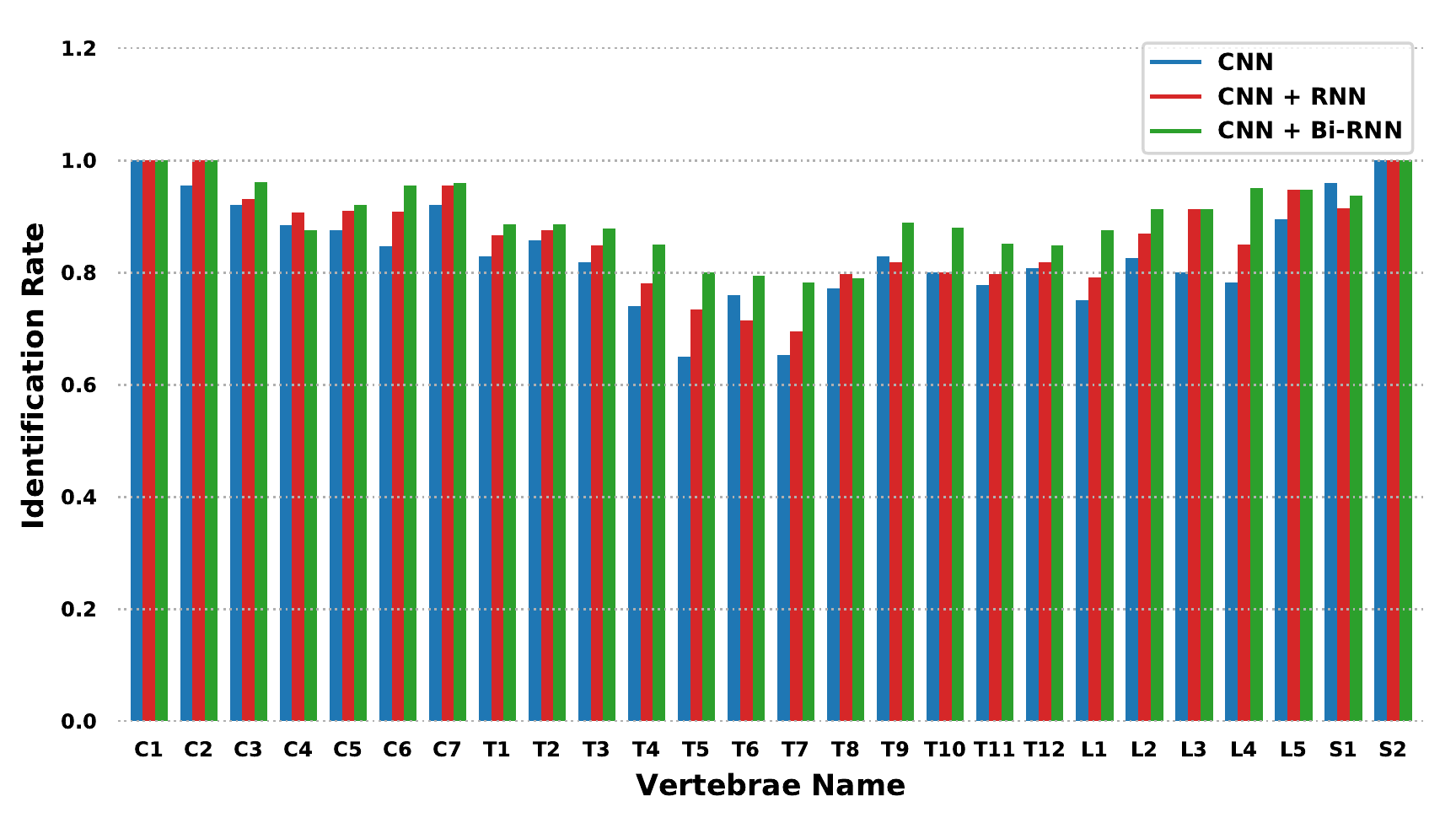}
\caption{}
\end{subfigure}
\begin{subfigure}[b]{0.5\textwidth}
\centering
\includegraphics[width=\linewidth]{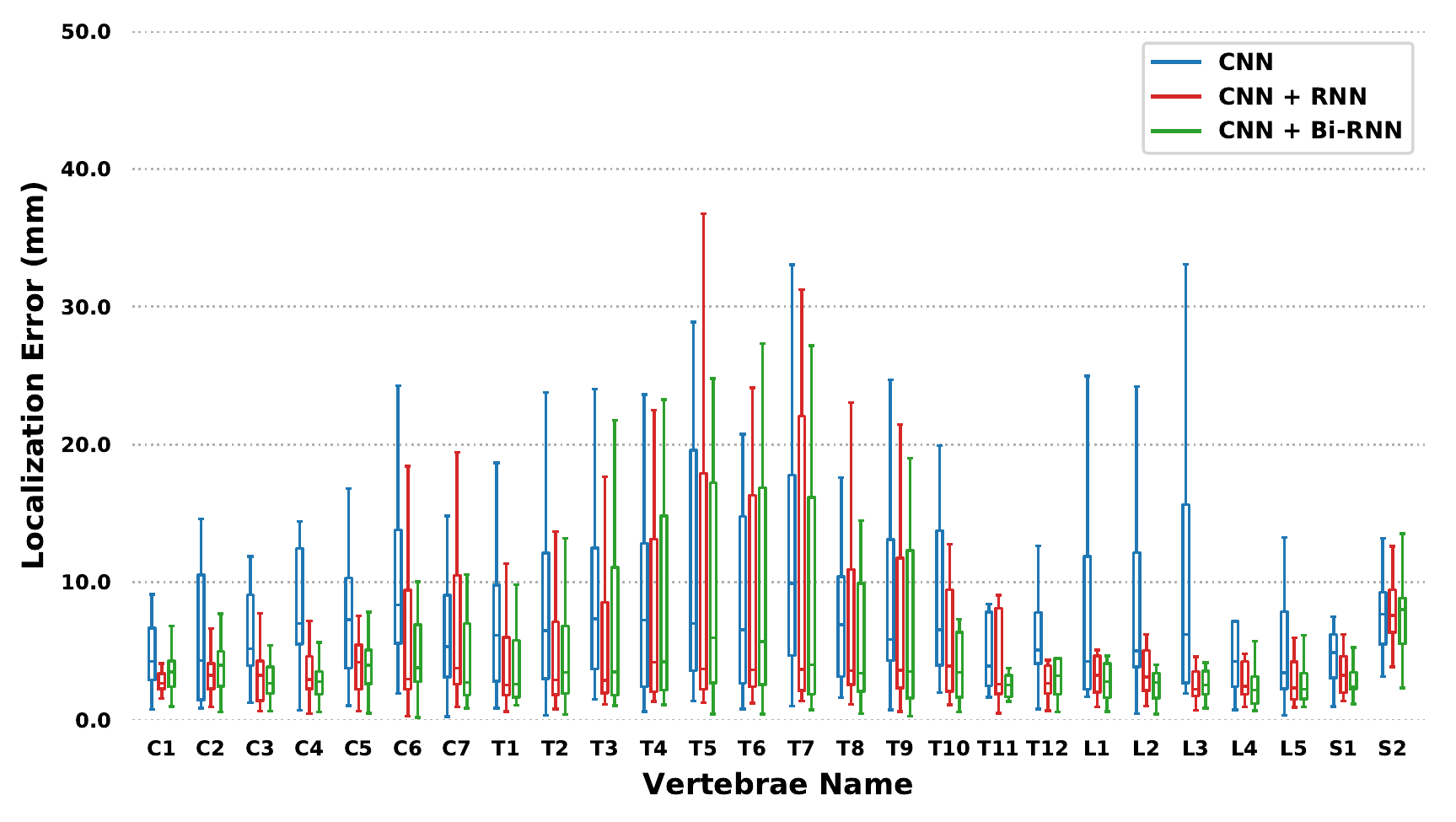}
\caption{}
\end{subfigure}
\caption{Vertebra-wise identification and localization results. Left: the identification accuracy of each vertebra; Right: the localization error statistics of each vertebra.}
\label{fig: results_stats}
\end{figure*}

The dataset used in all of the experiments is a public dataset from spineweb \footnote{http://spineweb.digitalimaginggroup.ca}. It is considered challenging due to the variety of pathological cases, arbitrary of field-of-view and the existence of artificial implants. For all the experiments, we use the official split for training and testing as did by other state-of-the-art methods \cite{glocker2013vertebrae,chen2015automatic,DBLP:journals/corr/YangXXHLZXPCTCM17}. In total, there are 302 CT scans in this dataset. 242 CT scans from 125 patients are used for training and the rest 60 CT scans are held out for testing. During preprocessing, CT images are resampled such that the resolutions along the longtitudinal, frontal and sagittal axes are 1.25mm, 1.0mm and 1.0mm, respectively. Vertebrae locations are first normalized according to the new resolution and then converted to the corresponding voxel locations of images or image samples. Each vertebra centroid location in the CT sans is annotated along with the corresponding vertebrae type. All the experiments are conducted using the TensorFlow \footnote{https://www.tensorflow.org} platform on two NVIDIA GTX Geforce 1070 GPUs.

\subsection{Performance of Deep Multi-Task 3D CNN}

To train and evaluate the performance of the proposed deep multi-task 3D CNN, we randomly crop samples with size $32 \times 112 \times 96$ from CT scans. During the sample generation, we make sure that all the vertebrae are evenly sampled and, on the average, there are about $40$ samples for each vertebra. The network are trained for about $15$ epochs with $\text{batch\_size}=24$, $\text{learning\_rate}=0.001$, $\text{weight\_decay}=0.0001$, $\text{momentum}=0.9$ and $\lambda=0.12$. The learning rate is reduced every $20000$ iteration by a factor of $0.4$. All the hyperparameters and $\lambda$ are chosen empirically with validation. For this study we are not interested in finding the best parameter settings for the model. In general, we find $\lambda = 0.12$ works better in a multi-task scenario and values close to $0.12$ give minor performance difference. We refer readers to \cite{yang2016deep} for better multi-task parameter choices. Two evaluation metrics are used: \textit{sample classification accuracy} and \textit{sample localization error}. Sample classification accuracy is the number of samples that are successfully classified among all the testing samples. Average sample localization error is defined as the average distances (in mm) between the predicted locations and vertebrae centroids
\begin{equation}
e = \frac{\sum_{i=1}^N||l^i_{pred}-l^i_{gt}||}{N}
\end{equation}
where $l^i_{pred}$ and $l^i_{gt}$ denote the predicted location and the ground truth centroid for the $i$-th positive sample, respectively.

To demonstrate the effectiveness of the proposed multi-task 3D CNN, we compare our approach with 3 other baseline methods: 2D CNN MTL, 3D CNN ID and 3D CNN LOC.
For 2D CNN MTL, we convert all the 3D convolutional/pooling layers of the proposed network to their 2D versions. For 3D CNN ID and 3D CNN LOC, we remove the identification loss and the localization loss, respectively. We train theses three methods using the similar hyperparameters as the 3D CNN MTL. The evaluation results are shown in Table \ref{tab: spine_exp_cnn}.

We can see that 2D CNN MTL only achieves $43.19\%$ classification accuracy and $8.74$ mm localization error which is much worse than its 3D counterpart. Since 2D convolution can not encode the important spatial information of 3D images, the degradation in performance is expected. The classification accuracy of 3D CNN ID is $48.52\%$ which is significantly better than 2D CNN MTL due to the use of 3D convolution. However, its performance is still worse than 3D CNN MTL, which achieves a $52.39\%$ classification accuracy. This demonstrates that training identification and localization jointly is very helpful in improving the network's ability to distinguish different vertebrae. The localization error of 3D CNN LOC and 3D CNN MTL are $7.05$ and $7.03$ mm, respectively. Such a close performance in localization error demonstrates that finding the vertebra centroids does not necessarily require recognizing vertebrae type which is consistent with common sense. However, the classification accuracy overall is not so good. This is because each cropped sample has a very narrow field-of-view that contains limited contextual information. Since different vertebrae are very similar in appearance, distinguishing between vertebrae, especially those neighboring ones, is very challenging without more contextual information.

\subsection{Overall Performance}

\begin{figure*}
  \centering
  \begin{subfigure}[b]{1.0\textwidth}
  \centering
  \includegraphics[width=0.93\linewidth]{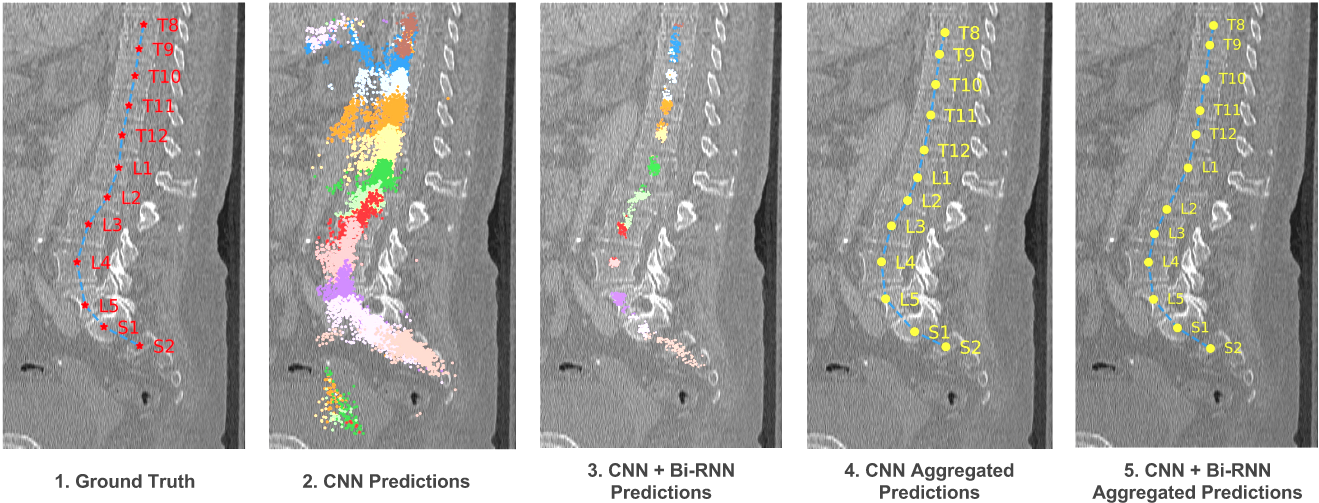}
  \caption{}
  \end{subfigure}
  \begin{subfigure}[b]{1.0\textwidth}
  \centering
  \includegraphics[width=0.93\linewidth]{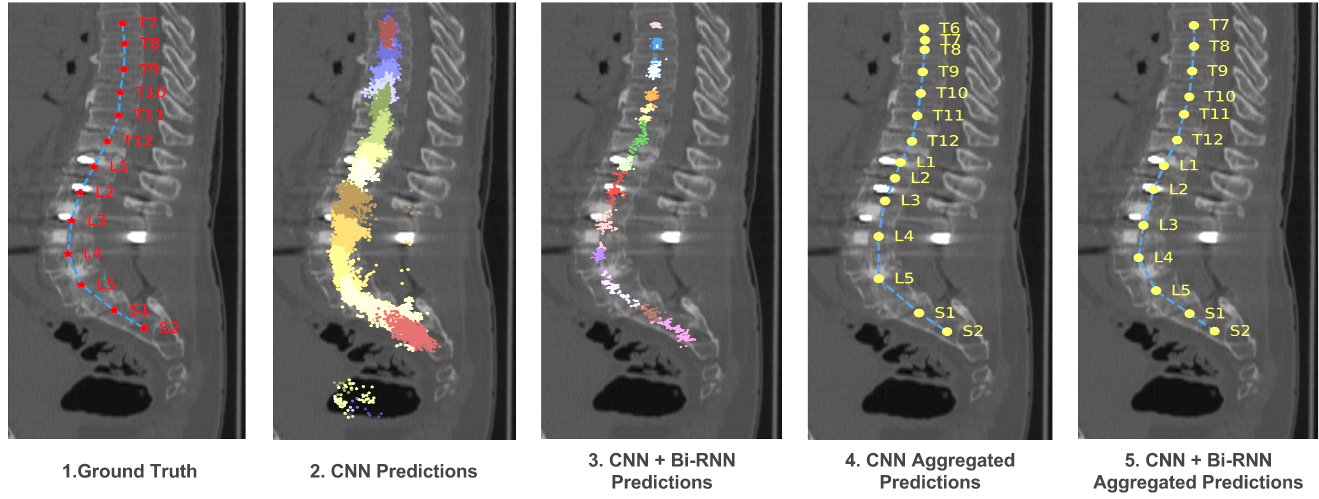}
  \caption{}
  \end{subfigure}
  \caption{Two  example successful cases of using the proposed method and the CNN only baseline method.}
  \label{fig: spine_success}
\end{figure*}

\begin{figure*}
  \centering
  \begin{subfigure}[b]{1.0\textwidth}
  \centering
  \includegraphics[width=0.93\linewidth]{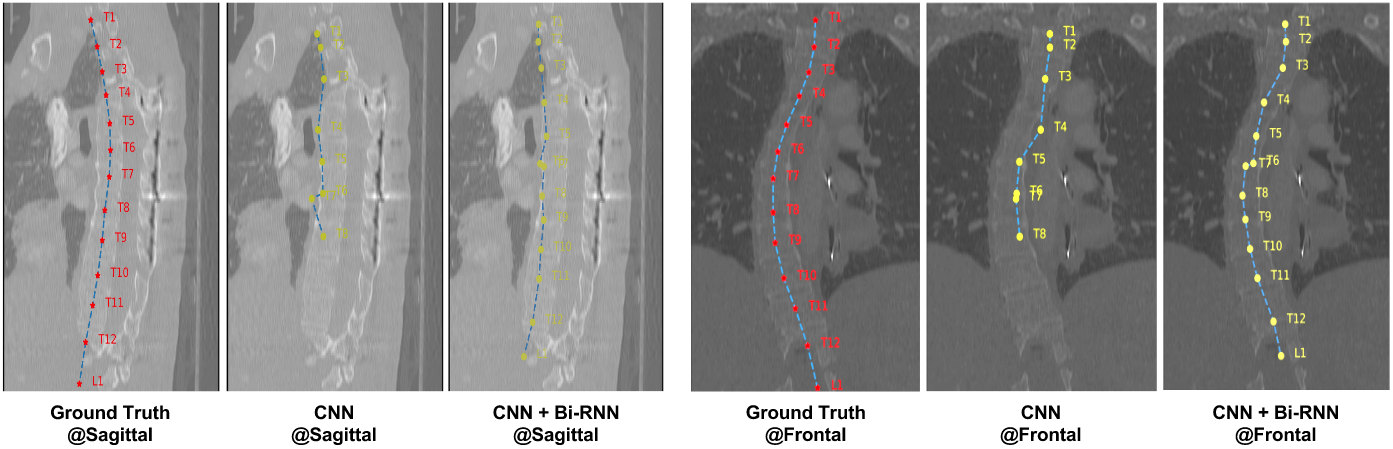}
  \vspace{-5mm}
  \caption{}
  \end{subfigure}
  \begin{subfigure}[b]{1.0\textwidth}
  \centering
  \includegraphics[width=0.93\linewidth]{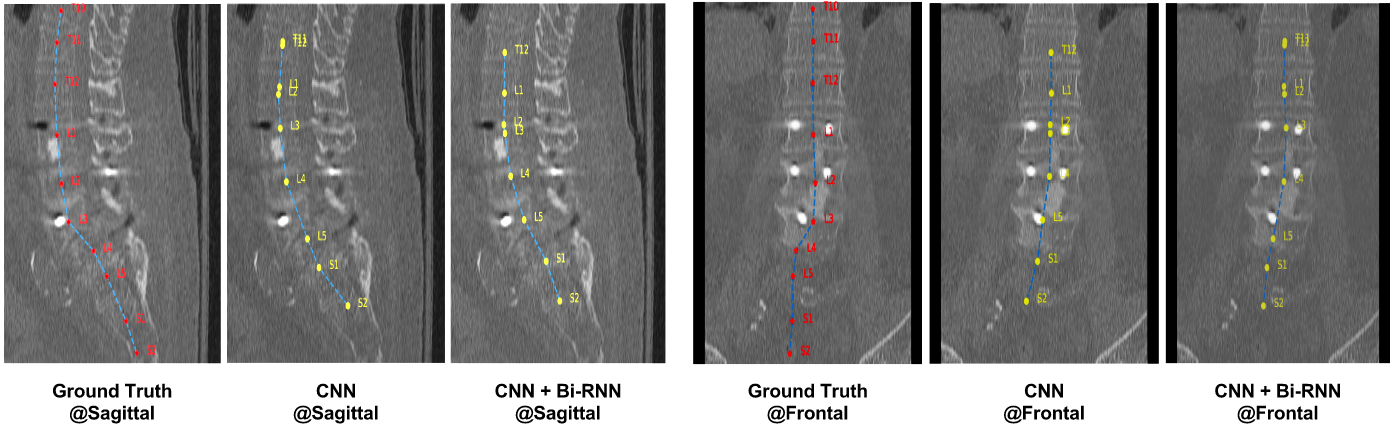}
  \caption{}
  \end{subfigure}
  \caption{Two example failure cases of using the proposed method and the CNN only baseline method.}
  \label{fig: spine_fail}
\end{figure*}

To train the multi-task Bi-RNN, we first randomly crop subimages of various sizes from the CT scans used for training. The number of cropped subimages is proportional to the number of vertebrae inside a CT scan. On the average, $30$ subimages are generated for each vertebra and in total, we obtain about $70,000$ subimages. Due to memory limitation, the maximum subimage size is $96 \times 256 \times 256$ which covers a maximum of $8$ vertebrae and gives longer enough contextual information. For each of the subimage, we generate a sequence of sample features in the way described in Section \ref{sec: spine_rnn}. The average sequence length $T$ is $266$. We train the Bi-RNN for about $12$ epochs with $\text{batch\_size}=256$, $\text{learning\_rate}=10^{-6}$, $\text{weight\_decay}=0.0001$, $\text{momentum}=0.9$ and $\lambda=0.10$. Again the hyperparameters are chosen empirically with validataion. The trained Bi-RNN in combination with the trained FCN is used to evaluate the testing CT scans. We use \textit{identification rate} and \textit{localization error} as the evaluation metrics following the definition from \cite{glocker2012automatic}.

The overall performance results of the proposed method are given in Table \ref{tab: spine_exp_rnn}. Here, we compare our method with three state-of-the-art methods on the same dataset. To demonstrate the effectiveness of the Bi-RNN, we also compare our method (denoted as CNN + Bi-RNN) with two baseline methods: 1) using CNN (denoted as CNN) only, 2) using CNN together with conventional RNN (denoted as CNN + RNN). For the CNN only baseline method, we use the trained FCN to generate dense predictions in a sliding window manner and the dense predictions for each vertebra will then be aggregated and refined to give the final identification and localization results. Table \ref{tab: spine_exp_rnn} shows the performance for all vertebrae types as well as the performance for each of the vertebrae categories (cervical, thoracic and lumbar). Both the mean and standard deviation of the localization errors are measured. We can see from Table \ref{tab: spine_exp_rnn} that the proposed method outperforms both the state-of-the-art methods and the baseline methods in most of the measurements. For the standard deviation of the localization errors, the method from Dong et al. \cite{DBLP:journals/corr/YangXXHLZXPCTCM17} gives similar performances with our method. This is because they used a message passing and sparsity regularization algorithm during the refinement step to suppress the outliers. This scheme can be also added to our method for further performance improvement. We then analyze the vertebra-wise performance of the proposed approach and compare it with the baseline methods. As shown in Figure \ref{fig: results_stats}, we can find that the proposed approach performs better than the baseline methods on most vertebrae. This demonstrates that using RNN in combination with CNN can give better long-range contextual understanding and yield better performance. We also find that using Bi-RNN against convensional RNN can further boost the performance. This is consistent with the observation that contextual information comes from two directions (below and above) and should be addressed accordingly.

\subsection{Success and Failure Cases}

Figure \ref{fig: spine_success} shows two successful identification and localization results. In the second and third columns, each colored point denotes the predicted vertebrae location for a sample. For images (a)2 and (b)2, the colored points are from the CNN only baseline method that samples the image using FCN in a sliding window manner. For images (a)3 and (b)3, the colored points are from the proposed method that makes predictions for each of the positive feature samples of the input sequence. As we can see here, the dense predictions from the proposed method are more concentrated around the vertebrae centroids which indicates more accurate prediction results. The fourth and fifth columns are the aggregated predictions from the dense predictions for the CNN only baseline method and the proposed method, respectively. Compared with the ground truths in the first column, we can see that both the proposed and CNN only baseline method perform well and the proposed method is slightly better than the CNN only baseline method.

Besides the success cases, we also investigate when the proposed method does not work. Figure \ref{fig: spine_fail} shows two challenging examples that the CNN only baseline method or proposed method fail. For both examples, we show the results of the sagittal and frontal view, respectively. Figure \ref{fig: spine_fail} (a) shows a pathological example with blurred vertebra boundaries. We can see that most of the predictions by the CNN only baseline method are incorrect. On the other hand, the predictions from the proposed method follow the spine structure and in general are acceptable. Figure \ref{fig: spine_fail} (b) is even more challenging and both the baseline and proposed methods fail.

\section{Conclusion}

We present a novel approach to vertebrae identification and localization from CT scans. Due to the similarity of vertebrae appearance and the variability of spinal images, such as arbitrary field-of-view, vertebrae curvatures,  we develop a data-driven learning-based method to robustly capture both the short-range and long-range contextual information that are critical for vertebrae identification and localization. For the short-range contextual information, we train an MTL 3D CNN that effectively extracts the features of vertebral samples by leveraging the domain information contained in both the vertebrae identification and localization tasks. The use of 3D convolutions enables it to encode 3D spatial information of CT volumes to yield a more robust model than the 2D counterparts. For the long-range contextual information, we develop a bidirectional MTL RNN that inherently learns the anatomic structure in a data-driven manner and exploits the contextual information among vertebral samples during testing phase. Experimental results demonstrate that the proposed MTL 3D CNN/FCN extracts better feature representations than its 2D or single-task counterparts,  outperforming the state-of-the-art on a challenging dataset by a significant margin.

% use section* for acknowledgment
\section*{Acknowledgment}

The authors would like to thank the support of the New York State through the Goergen Institute for Data Science, and the J. Robert Gladden Orthopaedic Society Award.

% Can use something like this to put references on a page
% by themselves when using endfloat and the captionsoff option.
\ifCLASSOPTIONcaptionsoff
  \newpage
\fi

% trigger a \newpage just before the given reference
% number - used to balance the columns on the last page
% adjust value as needed - may need to be readjusted if
% the document is modified later
%\IEEEtriggeratref{8}
% The "triggered" command can be changed if desired:
%\IEEEtriggercmd{\enlargethispage{-5in}}

% references section

% can use a bibliography generated by BibTeX as a .bbl file
% BibTeX documentation can be easily obtained at:
% http://mirror.ctan.org/biblio/bibtex/contrib/doc/
% The IEEEtran BibTeX style support page is at:
% http://www.michaelshell.org/tex/ieeetran/bibtex/
\bibliographystyle{IEEEtran}
% argument is your BibTeX string definitions and bibliography database(s)
\bibliography{IEEEabrv,references}

% Generated by IEEEtran.bst, version: 1.14 (2015/08/26)
\begin{thebibliography}{10}
\providecommand{\url}[1]{#1}
\csname url@samestyle\endcsname
\providecommand{\newblock}{\relax}
\providecommand{\bibinfo}[2]{#2}
\providecommand{\BIBentrySTDinterwordspacing}{\spaceskip=0pt\relax}
\providecommand{\BIBentryALTinterwordstretchfactor}{4}
\providecommand{\BIBentryALTinterwordspacing}{\spaceskip=\fontdimen2\font plus
\BIBentryALTinterwordstretchfactor\fontdimen3\font minus
  \fontdimen4\font\relax}
\providecommand{\BIBforeignlanguage}[2]{{%
\expandafter\ifx\csname l@#1\endcsname\relax
\typeout{** WARNING: IEEEtran.bst: No hyphenation pattern has been}%
\typeout{** loaded for the language `#1'. Using the pattern for}%
\typeout{** the default language instead.}%
\else
\language=\csname l@#1\endcsname
\fi
#2}}
\providecommand{\BIBdecl}{\relax}
\BIBdecl

\bibitem{Burns2015}
J.~E. Burns, \emph{Imaging of the Spine: A Medical and Physical
  Perspective}.\hskip 1em plus 0.5em minus 0.4em\relax Cham: Springer
  International Publishing, 2015, pp. 3--29.

\bibitem{ben2012vertebral}
I.~B. Ayed, K.~Punithakumar, R.~Minhas, R.~Joshi, and G.~J. Garvin, ``Vertebral
  body segmentation in mri via convex relaxation and distribution matching,''
  in \emph{International Conference on Medical Image Computing and
  Computer-Assisted Intervention}.\hskip 1em plus 0.5em minus 0.4em\relax
  Berlin, Heidelberg: Springer, 2012, pp. 520--527.

\bibitem{alomari2015vertebral}
R.~S. Alomari, S.~Ghosh, J.~Koh, and V.~Chaudhary, ``Vertebral column
  localization, labeling, and segmentation,'' in \emph{Spinal Imaging and Image
  Analysis}.\hskip 1em plus 0.5em minus 0.4em\relax Berlin, Heidelberg:
  Springer, 2015, pp. 193--229.

\bibitem{lecron2012fast}
F.~Lecron, J.~Boisvert, S.~Mahmoudi, H.~Labelle, and M.~Benjelloun, ``Fast 3d
  spine reconstruction of postoperative patients using a multilevel statistical
  model,'' in \emph{International Conference on Medical Image Computing and
  Computer-Assisted Intervention}.\hskip 1em plus 0.5em minus 0.4em\relax
  Berlin, Heidelberg: Springer, 2012, pp. 446--453.

\bibitem{simons2014fast}
C.~J. Simons, L.~Cobb, and B.~S. Davidson, ``A fast, accurate, and reliable
  reconstruction method of the lumbar spine vertebrae using positional mri,''
  \emph{Annals of biomedical engineering}, vol.~42, no.~4, pp. 833--842, 2014.

\bibitem{otake2012automatic}
Y.~Otake, S.~Schafer, J.~Stayman, W.~Zbijewski, G.~Kleinszig, R.~Graumann,
  A.~Khanna, and J.~Siewerdsen, ``Automatic localization of vertebral levels in
  x-ray fluoroscopy using 3d-2d registration: a tool to reduce wrong-site
  surgery,'' \emph{Physics in medicine and biology}, vol.~57, no.~17, p. 5485,
  2012.

\bibitem{ruizrobust}
O.~E. Ruiz and J.~Fl{\'o}rez, ``Robust ct to us 3d-3d registration by using
  principal component analysis and kalman filtering,'' in \emph{Computational
  Methods and Clinical Applications for Spine Imaging: Third International
  Workshop and Challenge, CSI 2015, Held in Conjunction with MICCAI 2015,
  Munich, Germany, October 5, 2015, Proceedings}, vol. 9402.\hskip 1em plus
  0.5em minus 0.4em\relax Berlin, Heidelberg: Springer, 2016, p.~52.

\bibitem{yao2012detection}
J.~Yao, J.~E. Burns, H.~Munoz, and R.~M. Summers, ``Detection of vertebral body
  fractures based on cortical shell unwrapping,'' in \emph{International
  Conference on Medical Image Computing and Computer-Assisted
  Intervention}.\hskip 1em plus 0.5em minus 0.4em\relax Berlin, Heidelberg:
  Springer, 2012, pp. 509--516.

\bibitem{al2013compression}
S.~Al-Helo, R.~S. Alomari, S.~Ghosh, V.~Chaudhary, G.~Dhillon, A.-Z. Moh'd~B,
  H.~Hiary, and T.~M. Hamtini, ``Compression fracture diagnosis in lumbar: a
  clinical cad system,'' \emph{International journal of computer assisted
  radiology and surgery}, vol.~8, no.~3, pp. 461--469, 2013.

\bibitem{wang2016detection}
Y.~Wang, J.~Yao, J.~E. Burns, J.~Liu, and R.~M. Summers, ``Detection of
  degenerative osteophytes of the spine on pet/ct using region-based
  convolutional neural networks,'' in \emph{International Workshop on
  Computational Methods and Clinical Applications for Spine Imaging}.\hskip 1em
  plus 0.5em minus 0.4em\relax Springer, 2016, pp. 116--124.

\bibitem{linte2015toward}
C.~A. Linte, K.~E. Augustine, J.~J. Camp, R.~A. Robb, and D.~R. Holmes~III,
  ``Toward virtual modeling and templating for enhanced spine surgery
  planning,'' in \emph{Spinal Imaging and Image Analysis}.\hskip 1em plus 0.5em
  minus 0.4em\relax Berlin, Heidelberg: Springer, 2015, pp. 441--467.

\bibitem{knez2016manual}
D.~Knez, J.~Mohar, R.~J. Cirman, B.~Likar, F.~Pernu{\v{s}}, and T.~Vrtovec,
  ``Manual and computer-assisted pedicle screw placement plans: A quantitative
  comparison,'' in \emph{International Workshop on Computational Methods and
  Clinical Applications for Spine Imaging}.\hskip 1em plus 0.5em minus
  0.4em\relax Springer, 2016, pp. 105--115.

\bibitem{kumar2015robotic}
R.~Kumar, ``Robotic assistance and intervention in spine surgery,'' in
  \emph{Spinal Imaging and Image Analysis}.\hskip 1em plus 0.5em minus
  0.4em\relax Berlin, Heidelberg: Springer, 2015, pp. 495--506.

\bibitem{ma2010hierarchical}
J.~Ma, L.~Lu, Y.~Zhan, X.~Zhou, M.~Salganicoff, and A.~Krishnan, ``Hierarchical
  segmentation and identification of thoracic vertebra using learning-based
  edge detection and coarse-to-fine deformable model,'' in \emph{International
  Conference on Medical Image Computing and Computer-Assisted
  Intervention}.\hskip 1em plus 0.5em minus 0.4em\relax Berlin, Heidelberg:
  Springer, 2010, pp. 19--27.

\bibitem{schmidt2007spine}
S.~Schmidt, J.~Kappes, M.~Bergtholdt, V.~Pekar, S.~Dries, D.~Bystrov, and
  C.~Schn{\"o}rr, ``Spine detection and labeling using a parts-based graphical
  model,'' in \emph{Information Processing in Medical Imaging}.\hskip 1em plus
  0.5em minus 0.4em\relax Berlin, Heidelberg: Springer, 2007, pp. 122--133.

\bibitem{kelm2010detection}
B.~M. Kelm, S.~K. Zhou, M.~Suehling, Y.~Zheng, M.~Wels, and D.~Comaniciu,
  ``Detection of 3d spinal geometry using iterated marginal space learning,''
  in \emph{International MICCAI Workshop on Medical Computer Vision}.\hskip 1em
  plus 0.5em minus 0.4em\relax Springer, 2010, pp. 96--105.

\bibitem{zhan2012robust}
Y.~Zhan, D.~Maneesh, M.~Harder, and X.~S. Zhou, ``Robust mr spine detection
  using hierarchical learning and local articulated model,'' in
  \emph{International Conference on Medical Image Computing and
  Computer-Assisted Intervention}.\hskip 1em plus 0.5em minus 0.4em\relax
  Berlin, Heidelberg: Springer, 2012, pp. 141--148.

\bibitem{glocker2012automatic}
B.~Glocker, J.~Feulner, A.~Criminisi, D.~R. Haynor, and E.~Konukoglu,
  ``Automatic localization and identification of vertebrae in arbitrary
  field-of-view ct scans,'' in \emph{International Conference on Medical Image
  Computing and Computer-Assisted Intervention}.\hskip 1em plus 0.5em minus
  0.4em\relax Berlin, Heidelberg: Springer, 2012, pp. 590--598.

\bibitem{glocker2013vertebrae}
B.~Glocker, D.~Zikic, E.~Konukoglu, D.~R. Haynor, and A.~Criminisi, ``Vertebrae
  localization in pathological spine ct via dense classification from sparse
  annotations,'' in \emph{International Conference on Medical Image Computing
  and Computer-Assisted Intervention}.\hskip 1em plus 0.5em minus 0.4em\relax
  Springer, 2013, pp. 262--270.

\bibitem{chen2015automatic}
H.~Chen, C.~Shen, J.~Qin, D.~Ni, L.~Shi, J.~C. Cheng, and P.-A. Heng,
  ``Automatic localization and identification of vertebrae in spine ct via a
  joint learning model with deep neural networks,'' in \emph{International
  Conference on Medical Image Computing and Computer-Assisted
  Intervention}.\hskip 1em plus 0.5em minus 0.4em\relax Springer, 2015, pp.
  515--522.

\bibitem{DBLP:journals/corr/HarrisonXGLSM17}
A.~P. Harrison, Z.~Xu, K.~George, L.~Lu, R.~M. Summers, and D.~J. Mollura,
  ``Progressive and multi-path holistically nested neural networks for
  pathological lung segmentation from ct images,'' in \emph{Medical Image
  Computing and Computer-Assisted Intervention − MICCAI 2017}, M.~Descoteaux,
  L.~Maier-Hein, A.~Franz, P.~Jannin, D.~L. Collins, and S.~Duchesne,
  Eds.\hskip 1em plus 0.5em minus 0.4em\relax Cham: Springer International
  Publishing, 2017, pp. 621--629.

\bibitem{cai2016pancreas}
J.~Cai, L.~Lu, Z.~Zhang, F.~Xing, L.~Yang, and Q.~Yin, ``Pancreas segmentation
  in mri using graph-based decision fusion on convolutional neural networks,''
  in \emph{International Conference on Medical Image Computing and
  Computer-Assisted Intervention}.\hskip 1em plus 0.5em minus 0.4em\relax
  Springer, 2016, pp. 442--450.

\bibitem{dou2016automatic}
Q.~Dou, H.~Chen, L.~Yu, L.~Zhao, J.~Qin, D.~Wang, V.~C. Mok, L.~Shi, and P.-A.
  Heng, ``Automatic detection of cerebral microbleeds from mr images via 3d
  convolutional neural networks,'' \emph{IEEE transactions on medical imaging},
  vol.~35, no.~5, pp. 1182--1195, 2016.

\bibitem{DBLP:journals/corr/YangXXHLZXPCTCM17}
D.~Yang, T.~Xiong, D.~Xu, Q.~Huang, D.~Liu, S.~K. Zhou, Z.~Xu, J.~Park,
  M.~Chen, T.~D. Tran, S.~P. Chin, D.~Metaxas, and D.~Comaniciu, ``Automatic
  vertebra labeling in large-scale 3d ct using deep image-to-image network with
  message passing and sparsity regularization,'' in \emph{Information
  Processing in Medical Imaging}, M.~Niethammer, M.~Styner, S.~Aylward, H.~Zhu,
  I.~Oguz, P.-T. Yap, and D.~Shen, Eds.\hskip 1em plus 0.5em minus 0.4em\relax
  Cham: Springer International Publishing, 2017, pp. 633--644.

\bibitem{ronneberger2015u}
O.~Ronneberger, P.~Fischer, and T.~Brox, ``U-net: Convolutional networks for
  biomedical image segmentation,'' in \emph{International Conference on Medical
  Image Computing and Computer-Assisted Intervention}.\hskip 1em plus 0.5em
  minus 0.4em\relax Springer, 2015, pp. 234--241.

\bibitem{girshick2015fast}
R.~Girshick, ``Fast r-cnn,'' in \emph{Proceedings of the IEEE International
  Conference on Computer Vision}, 2015, pp. 1440--1448.

\bibitem{chen2015automaticfetal}
H.~Chen, Q.~Dou, D.~Ni, J.-Z. Cheng, J.~Qin, S.~Li, and P.-A. Heng, ``Automatic
  fetal ultrasound standard plane detection using knowledge transferred
  recurrent neural networks,'' in \emph{International Conference on Medical
  Image Computing and Computer-Assisted Intervention}.\hskip 1em plus 0.5em
  minus 0.4em\relax Springer, 2015, pp. 507--514.

\bibitem{kong2016recognizing}
B.~Kong, Y.~Zhan, M.~Shin, T.~Denny, and S.~Zhang, ``Recognizing end-diastole
  and end-systole frames via deep temporal regression network,'' in
  \emph{International Conference on Medical Image Computing and
  Computer-Assisted Intervention}.\hskip 1em plus 0.5em minus 0.4em\relax
  Springer, 2016, pp. 264--272.

\bibitem{chen2016combining}
J.~Chen, L.~Yang, Y.~Zhang, M.~Alber, and D.~Z. Chen, ``Combining fully
  convolutional and recurrent neural networks for 3d biomedical image
  segmentation,'' in \emph{Advances in Neural Information Processing Systems},
  2016, pp. 3036--3044.

\bibitem{DBLP:journals/corr/CaiLXXY17}
J.~Cai, L.~Lu, Y.~Xie, F.~Xing, and L.~Yang, ``Improving deep pancreas
  segmentation in {CT} and {MRI} images via recurrent neural contextual
  learning and direct loss function,'' in \emph{Medical Image Computing and
  Computer-Assisted Intervention − MICCAI 2017}, M.~Descoteaux,
  L.~Maier-Hein, A.~Franz, P.~Jannin, D.~L. Collins, and S.~Duchesne,
  Eds.\hskip 1em plus 0.5em minus 0.4em\relax Cham: Springer International
  Publishing, 2017, pp. 674–--682.

\bibitem{suzani2015fast}
A.~Suzani, A.~Seitel, Y.~Liu, S.~Fels, R.~N. Rohling, and P.~Abolmaesumi,
  ``Fast automatic vertebrae detection and localization in pathological ct
  scans-a deep learning approach,'' in \emph{International Conference on
  Medical Image Computing and Computer-Assisted Intervention}.\hskip 1em plus
  0.5em minus 0.4em\relax Springer, 2015, pp. 678--686.

\bibitem{sermanet2013overfeat}
P.~Sermanet, D.~Eigen, X.~Zhang, M.~Mathieu, R.~Fergus, and Y.~LeCun,
  ``Overfeat: Integrated recognition, localization and detection using
  convolutional networks,'' \emph{arXiv preprint arXiv:1312.6229}, 2013.

\bibitem{long2015fully}
J.~Long, E.~Shelhamer, and T.~Darrell, ``Fully convolutional networks for
  semantic segmentation,'' in \emph{Proceedings of the IEEE Conference on
  Computer Vision and Pattern Recognition}, 2015, pp. 3431--3440.

\bibitem{schuster1997bidirectional}
M.~Schuster and K.~K. Paliwal, ``Bidirectional recurrent neural networks,''
  \emph{IEEE Transactions on Signal Processing}, vol.~45, no.~11, pp.
  2673--2681, 1997.

\bibitem{krizhevsky2012imagenet}
A.~Krizhevsky, I.~Sutskever, and G.~E. Hinton, ``Imagenet classification with
  deep convolutional neural networks,'' in \emph{Advances in neural information
  processing systems}, 2012, pp. 1097--1105.

\bibitem{he2016deep}
K.~He, X.~Zhang, S.~Ren, and J.~Sun, ``Deep residual learning for image
  recognition,'' in \emph{Proceedings of the IEEE Conference on Computer Vision
  and Pattern Recognition}, 2016, pp. 770--778.

\bibitem{hochreiter1997long}
S.~Hochreiter and J.~Schmidhuber, ``Long short-term memory,'' \emph{Neural
  computation}, vol.~9, no.~8, pp. 1735--1780, 1997.

\bibitem{yang2016deep}
Y.~Yang and T.~Hospedales, ``Deep multi-task representation learning: A tensor
  factorisation approach,'' \emph{arXiv preprint arXiv:1605.06391}, 2016.

\end{thebibliography}

% that's all folks
\end{document}